\pdfoutput=1

\documentclass[11pt]{article}

\usepackage{emnlp2021}

\usepackage{times}
\usepackage{latexsym}

\usepackage[T1]{fontenc}

\usepackage[utf8]{inputenc}

\usepackage{microtype}

\usepackage{amsmath}
\usepackage{comment}
\usepackage{graphicx}
\usepackage{here}
\usepackage{qtree}
\usepackage{color}
\usepackage{booktabs,multirow}
\title{Modeling Human Sentence Processing \\with Left-Corner Recurrent Neural Network Grammars}

\author{Ryo Yoshida$^{1}$ {\rm and} Hiroshi Noji$^{2}$ {\rm and} Yohei Oseki$^{1}$\\
 $^1$The University of Tokyo\\
 $^2$Artificial Intelligence Research Center, AIST\\
\texttt{\{yoshiryo0617, oseki\}@g.ecc.u-tokyo.ac.jp}\\
\texttt{hiroshi.noji@aist.go.jp}}

\begin{document}
\maketitle

\begin{abstract}
In computational linguistics, it has been shown that hierarchical structures make language models (LMs) more human-like. However, the previous literature has been agnostic about a parsing strategy of the hierarchical models. In this paper, we investigated whether hierarchical structures make LMs more human-like, and if so, which parsing strategy is most cognitively plausible. In order to address this question, we evaluated three LMs against human reading times in Japanese with head-final left-branching structures: Long Short-Term Memory (LSTM) as a sequential model and Recurrent Neural Network Grammars (RNNGs) with top-down and left-corner parsing strategies as hierarchical models. Our computational modeling demonstrated that left-corner RNNGs outperformed top-down RNNGs and LSTM, suggesting that hierarchical and left-corner architectures are more cognitively plausible than top-down or sequential architectures. In addition, the relationships between the cognitive plausibility and (i) perplexity, (ii) parsing, and (iii) beam size will also be discussed.\footnote{Our codes are available at \url{https://github.com/osekilab/RNNG-EyeTrack}.}
\end{abstract}

\section{Introduction}
\label{sec:intro}
It has been debated in computational linguistics whether language models (LMs) become more human-like by explicitly modeling hierarchical structures of natural languages. Previous work has revealed that while sequential models such as recurrent neural networks (RNNs) can successfully predict human reading times \citep{frankInsensitivityHumanSentenceProcessing2011}, there is an advantage in explicitly modeling hierarchical structures \citep{fossumSequentialVsHierarchical2012a}. More recently, RNNs that explicitly model hierarchical structures, namely Recurrent Neural Network Grammars (RNNGs, \citealp{dyerRecurrentNeuralNetwork2016a}), have attracted considerable attention, effectively capturing grammatical dependencies (e.g., subject-verb agreement) much better than RNNs in targeted syntactic evaluations \citep{kuncoroLSTMsCanLearn2018, wilcox-etal-2019-structural}. In addition, \citet{haleFindingSyntaxHuman2018} showed that RNNGs can successfully predict human brain activities, and recommended RNNGs as ``a mechanistic model of the syntactic processing that occurs during normal human sentence processing.''

However, this debate has focused almost exclusively on the dichotomy between the hierarchical and sequential models, without reference to the parsing strategies among the hierarchical models. Specifically, although \citet{dyerRecurrentNeuralNetwork2016a} and \citet{haleFindingSyntaxHuman2018} adopted the vanilla RNNG with a top-down parsing strategy for English with head-initial right-branching structures, \citet{abneyMemoryRequirementsLocal1991} and \citet{resnikLeftcornerParsingPsychological1992} suggested that the top-down parsing strategy is not optimal for head-final left-branching structures, and alternatively proposed the left-corner parsing strategy as more human-like parsing strategy.

In this paper, we investigate whether hierarchical structures make LMs more human-like, and if so, which parsing strategy is most cognitively plausible. In order to address this question, we evaluate three LMs against human reading times in Japanese with head-final left-branching structures: Long Short-Term Memory (LSTM) as a sequential model and Recurrent Neural Network Grammars (RNNGs) with top-down and left-corner parsing strategies as hierarchical models.

\section{Linking hypothesis}
\label{sec:linkhypo}
It is well established in psycholinguistics that humans predict next segments during sentence processing, and the less predictable the segment is, the more surprising that segment is. Surprisal theory \citep{haleProbabilisticEarleyParser2001, levyExpectationbasedSyntacticComprehension2008} quantifies this predictability of the segment as $-\log{p({\rm segment}| {\rm context})}$, an information-theoretic complexity metric known as \emph{surprisal}. In line with the previous literature (e.g., \citealp{SMITH2013302}), we employed this metric to logarithmically link probabilities estimated from LMs with cognitive efforts measured from humans. Intuitively, the cognitively plausible LMs will compute surprisals with similar trends as human cognitive efforts. Computational models of human sentence processing have been explored by comparing surprisals from various LMs with reading times (e.g., \citealp{frankInsensitivityHumanSentenceProcessing2011}) and brain activities (e.g., \citealp{frankERPResponseAmount2015}).

\section{Methods}
\label{sec:methods}
\subsection{Language models}
\label{subsec:methods}
\paragraph{Long Short-Term Memory (LSTM):}
LSTMs are a sequential model that does not model hierarchical structures. We used a 2-layer LSTM with 256 hidden and input dimensions. The implementation by \citet{gulordavaColorlessGreenRecurrent2018} was employed.


\paragraph{Recurrent Neural Network Grammar (RNNG):}
RNNGs are a hierarchical model that explicitly models hierarchical structures. We experimented with two types of stack-only RNNGs \citep{kuncoroWhatRecurrentNeural2017}: top-down RNNGs and left-corner RNNGs \citep{kuncoroLSTMsCanLearn2018}.\footnote{\citet{resnikLeftcornerParsingPsychological1992} suggested that an \emph{arc-eager} left-corner parsing strategy is cognitively plausible. \citet{jinMemoryboundedNeuralIncremental2020} implemented an incremental neural parser that builds tree structures with the arc-eager left-corner parsing strategy, but it requires an extremely large beam size to achieve the reasonable parsing accuracy. Thus, in this paper, we employed \emph{arc-standard} left-corner RNNGs as an approximation to the arc-eager left-corner parsing strategy that delayed attachments \citep{kuncoroLSTMsCanLearn2018}.} We used RNNGs that had a 2-layer stack LSTM with 256 hidden and input dimensions. The implementation by \citet{noji-oseki-2021-effective} was employed. Word-synchronous beam search \citep{sternEffectiveInferenceGenerative2017} was used for inference. RNNGs were evaluated with six beam sizes $k=\{100, 200, 400, 600, 800, 1000\}$.\footnote{$k$ means the action beam size. We set the word beam size to $k/10$ and the fast-track to $k/100$ \citep{sternEffectiveInferenceGenerative2017}.}


\subsection{Data sets}
\label{subsec:dat}
\paragraph{Training data:}
All LMs were trained on the National Institute for Japanese Language and Linguistics Parsed Corpus of Modern Japanese (NPCMJ), that comprises 67,018 sentences annotated with tree structures.\footnote{\url{http://npcmj.ninjal.ac.jp}} The sentences were split into subwords by a byte-pair encoding \citep{sennrichNeuralMachineTranslation2016}. LSTM used only terminal subwords, while RNNGs used terminal subwords and tree structures, both of which were trained sentence-level for 40 epochs and 3 times with different random seeds.\footnote{Following \citet{frankInsensitivityHumanSentenceProcessing2011}, traces and semantic information were removed in the way described in \citet{manningFoundationsStatisticalNatural1999}.}


\paragraph{Reading time data:}
All LMs were evaluated against first pass reading times from BCCWJ-EyeTrack \citep{asaharaReadingTimeAnnotationsBalanced2016}, that comprises 218 sentences annotated with eye-tracking reading times at each phrasal unit. Following \citet{asaharaReadingTimeAnnotationsBalanced2016}, the data points (i) not corresponding to the main text or (ii) not fixated were removed. In addition, following \citet{fossumSequentialVsHierarchical2012a}, the data points that contained subwords ``unknown'' to the LMs were also removed. Consequently, we included 12,114 data points in the statistical analyses among 19,717 data points in total.

\subsection{Evaluation metrics}
\label{subsec:eval}
\paragraph{Psychometric predictive power:}
We evaluated how well surprisal ($-\log{p({\rm segment}| {\rm context})}$) from each LM could predict human reading times. LMs process the sentences subword-by-subword, while reading times are annotated phrase-by-phrase. Thus, following \citet{wilcoxPredictivePowerNeural2020a}, the phrasal surprisal $I(p)$ was computed as the cumulative sum of surprisals of its constituent subwords $w_{l}, w_{l+1},\cdots,w_{m}$:
\begin{align}
    I(p) &= I(w_{l},w_{l+1},\cdots,w_{m}) =\sum_{i=l}^{m}I(w_{i})
    \label{eq:bunsurp}
\end{align}
where $I(w)$ is the surprisal of subword $w$:
\begin{align}
    I(w_{i}) &= -\log p(w_{i}|w_{<i})
    \label{eq:subsurp}
\end{align}

For the statistical analyses, we first trained a baseline regression model with several predictors that are known to affect reading times. Then, we added surprisal estimated from each LM as a predictor and evaluated the decrease in deviance ($\Delta D(LM)$) as psychometric predictive power:\footnote{In the previous literature, surprisals of the previous segments were also added as a predictor to capture spillover effects \citep{SMITH2013302}. In our experiments, we did not add surprisals of the previous segments because preliminary experiments showed that they were not significant for modeling reading times in all the LMs.}
\begin{align}
    \Delta D(LM) &= D_{B} - D_{LM}
    \label{eq:dev}
\end{align}
where $D_{B}$ and $D_{LM}$ are deviance of the baseline regression model and the regression model with surprisal, respectively. The details of our regression models are shown in Appendix~\ref{app:variables}.

\paragraph{Perplexity and parsing accuracy:}
\citet{goodkindPredictivePowerWord2018} demonstrated that perplexity of LMs and their psychometric predictive power are highly correlated. In order to investigate whether this correlation can be observed, perplexities of LMs were calculated based on the sentences in BCCWJ-EyeTrack.

In addition, given that RNNGs also serve as a parser, the correlation between parsing accuracy and psychometric predictive power was also investigated. The evaluation metric of parsing accuracy was the labeled bracketing F1. For this purpose, we used the sentences in NPCMJ because the sentences in BCCWJ-EyeTrack are not annotated with tree structures. Parsing accuracies of RNNGs were calculated based on the tree structures at the top of the final beam in word-synchronous beam search.

\section{Results and discussion}
\label{sec:result}
The result of our computational modeling is summarized in Figure~\ref{fig:ppl-dev_bccwj}: psychometric predictive power (the vertical axis) is plotted against perplexity (the horizontal axis).\footnote{The lower perplexity means the better next-subword prediction accuracy.} In this section, we first analyze psychometric predictive power itself, and then discuss its relationships with (i) perplexity, (ii) parsing, and (iii) beam size.

\subsection{Psychometric predictive power}
\label{subsec:cogacc}
Figure~\ref{fig:ppl-dev_bccwj} demonstrates that the hierarchical models (top-down/left-corner RNNGs) achieved higher psychometric predictive power than the sequential model (LSTM) and, among the hierarchical models, left-corner RNNGs achieved higher psychometric predictive power than top-down RNNGs. In order to confirm that these differences are statistically meaningful, we performed nested model comparisons. The result of nested model comparisons is summarized in Table~\ref{tbl:nestedcomp}, where the best result from each LM was compared.\footnote{Top-down RNNGs achieved the best result with the action beam size $1000$, while left-corner RNNGs achieved the best result with the action beam size $400$.} The significance threshold at $\alpha = 0.0056$ was imposed by the Bonferroni correction motivated by 9 tests ($0.05/9$).

First, nested model comparisons revealed that psychometric predictive power was significant for all LMs relative to the baseline regression model. The point here is that surprisals computed by LMs do explain human reading times in Japanese, generalizing the previous results in English.

Second, the hierarchical models (top-down/left-corner RNNGs) significantly outperformed the sequential model (LSTM), and the sequential model did not account for unique variances that the hierarchical models cannot explain. This result aligns with the previous observation that LMs become more human-like by explicitly modeling hierarchical structures on top of linear strings \citep{kuncoroLSTMsCanLearn2018, wilcox-etal-2019-structural, haleFindingSyntaxHuman2018}.

Finally, among the hierarchical models, left-corner RNNGs significantly outperformed top-down RNNGs, and top-down RNNGs did not account for unique variances that left-corner RNNGs cannot explain. This result corroborates \citet{abneyMemoryRequirementsLocal1991} from an information-theoretic perspective: the left-corner parsing strategy is more cognitively plausible than the top-down and bottom-up paring strategies.

Here we can conclude from these results that LMs become more human-like by explicitly modeling hierarchical structures and, most importantly, the left-corner parsing strategy was more cognitively plausible than the top-down parsing strategy.

{\setlength\textfloatsep{0pt}
\begin{figure}[t]
    \centering
      \includegraphics[width=7.5cm]{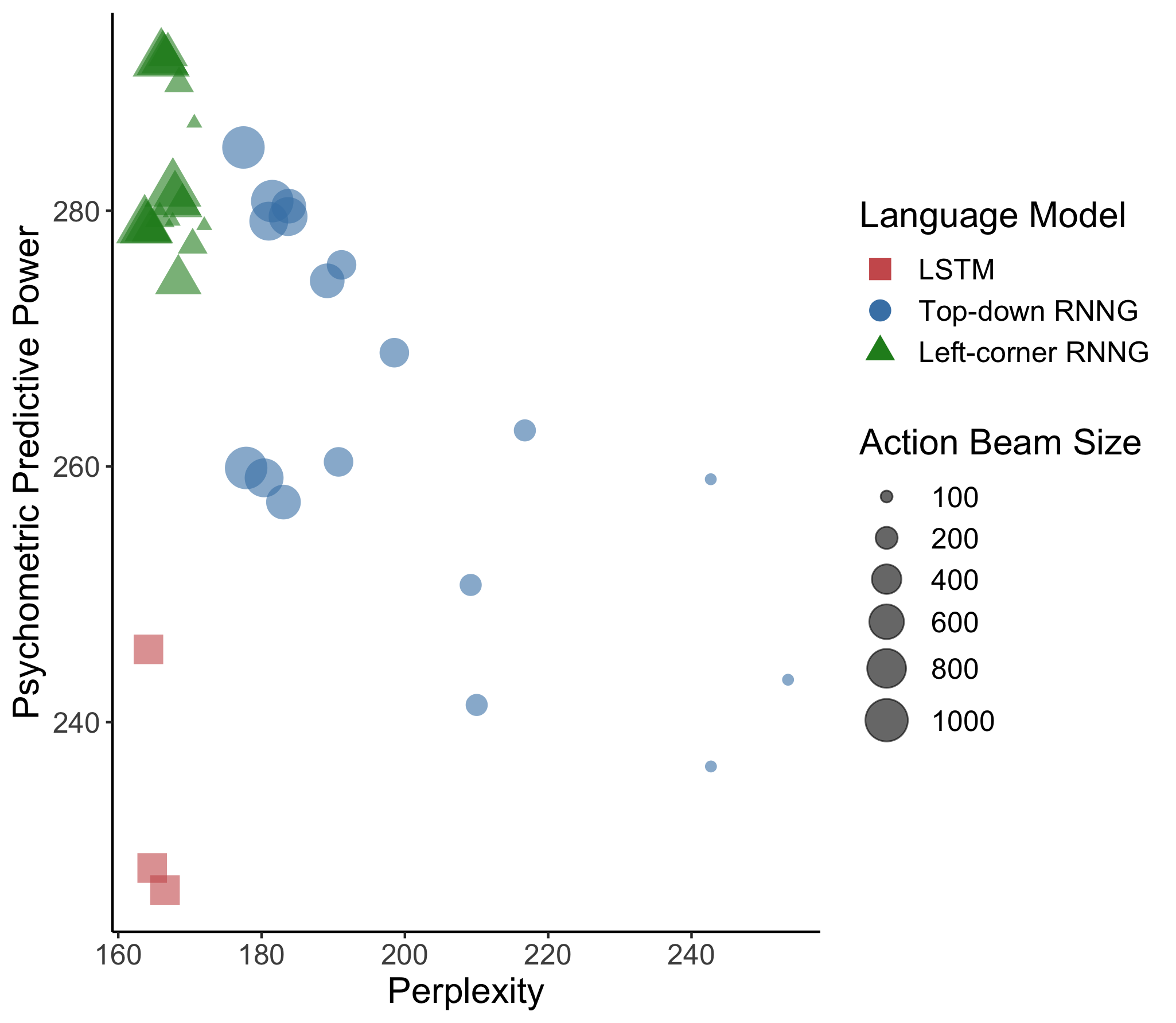}
      \vspace{-0.2cm}
      \caption{The result of our computational modeling: psychometric predictive power (the vertical axis) is plotted against perplexity (the horizontal axis).}
      \label{fig:ppl-dev_bccwj}
\end{figure}
}

\begin{table}
\centering
\begin{tabular}{lccc}
\toprule
\      & $\chi^2$ & df & $p$\\
\hline
Baseline < LSTM & 260.26 & 1 & <\textbf{0.0001}\\
Baseline < TD & 299.5 & 1 & <\textbf{0.0001}\\
Baseline < LC & 306.78 & 1 & <\textbf{0.0001}\\
LSTM < TD & 42.183 & 1 & <\textbf{0.0001}\\
LSTM < LC & 47.229 & 1 & <\textbf{0.0001}\\
TD < LSTM & 2.9406 & 1 & 0.08638\\
TD < LC & 11.84 & 1 & <\textbf{0.001}\\
LC < LSTM & 0.708 & 1 & 0.4001\\
LC < TD & 4.5609 & 1 & 0.03271\\
\bottomrule
\end{tabular}
\caption{The result of nested model comparisons with the best result from each LM. TD and LC indicate top-down and left-corner RNNGs, respectively. The significance threshold at $\alpha = 0.0056$ was imposed by the Bonferroni correction motivated by 9 tests ($0.05/9$).}
\label{tbl:nestedcomp}
\end{table}

\subsection{Perplexity}
\label{subsec:pplcogacc}
In this subsection, we discuss the relationship between perplexity and psychometric predictive power. First, Figure~\ref{fig:ppl-dev_bccwj} indicates that, among the hierarchical models, left-corner RNNGs, which achieved higher psychometric predictive power, also achieved lower perplexity than top-down RNNGs. Overall, the correlation between perplexity and psychometric predictive power of the hierarchical models was robust: the lower perplexity RNNGs have, the higher psychometric predictive power they also have. In sharp contrast, the correlation did not hold for the sequential model, where LSTMs achieved better perplexity, but worse psychometric predictive power than RNNGs with similar or even worse perplexity, corroborating \citet{goodkindPredictivePowerWord2018} that LSTM stands out as an outlier of the correlation between perplexity and psychometric predictive power.


\citet{kuribayashi-etal-2021-lower} recently showed that the correlation between perplexity and psychometric predictive power cannot be generalized to Japanese. They proposed that LMs are trained to flatten information density and thus satisfy the Uniform Information Density (UID) assumption \citep{genzel-charniak-2002-entropy, levy2005probabilistic,NIPS2006_c6a01432}, but information density in Japanese turned out not to be empirically uniform and far from the idealized UID assumption. At first, this proposal appears to be inconsistent with our results, but notice that \citet{kuribayashi-etal-2021-lower} only tested sequential models. Here we would like to suggest that, unlike sequential models, hierarchical models can be trained to be human-like, even in languages far from the idealized UID assumption.

\subsection{Parsing}
\label{subsec:f1cogacc}
In this subsection, we discuss the relationship between parsing accuracy and psychometric predictive power, which is summarized in Appendix~\ref{app:f1}, where psychometric predictive power (the vertical axis) is plotted against parsing accuracy (the horizontal axis). Interestingly, just like perplexity, left-corner RNNGs, which achieved higher psychometric predictive power, also achieved higher parsing accuracy than top-down RNNGs. Here again, the correlation between parsing accuracy and psychometric predictive power of the hierarchical models was robust: the higher parsing accuracy RNNGs have, the higher psychometric predictive power they also have.

\subsection{Beam size}
\label{subsec:beamcogacc}
Finally, we discuss the relationship between beam size and psychometric predictive power. The important generalization here is that, although top-down RNNGs improved in psychometric predictive power, perplexity, and parsing accuracy only when the beam size increased, left-corner RNNGs consistently performed well even with a small beam size. We interpret this generalization as demonstrating that left-corner RNNGs may be more human-like than top-down RNNGs in that they can infer the correct tree structure even with a small beam size comparable to humans. In order to reinforce this reasoning, we discuss (i) why left-corner RNNGs can infer the correct tree structure with a small beam size, and (ii) why inference with a small beam size is comparable to humans.

First, we show why left-corner RNNGs can infer the correct tree structure with a small beam size. Consider the following left-branching structures:
\vspace{0.2cm}

\qtreecentertrue
\noindent
a.\Tree[.{\textcolor{blue}{1}~X~\textcolor{white}{1}}
        [.{\textcolor{blue}{2}~X~\textcolor{white}{8}}
            [.{\textcolor{blue}{3}}~X~{\textcolor{white}{7}} 
                {\textcolor{blue}{4}~\emph{a}\textcolor{white}{11}}
                {\textcolor{blue}{5}~\emph{b}}
            ]
        {\textcolor{blue}{6}~\emph{c}}
        ]
        {\textcolor{blue}{7}~\emph{d}}
    ]
b.\Tree[.{~\textcolor{blue}{6}~X~\textcolor{white}{1}}
        [.{\textcolor{blue}{4}~X~\textcolor{white}{8}}
            [.{\textcolor{blue}{2}}~X~{\textcolor{white}{7}} 
                {\textcolor{blue}{1}~\emph{a}\textcolor{white}{11}}
                {\textcolor{blue}{3}~\emph{b}}
            ]
        {\textcolor{blue}{5}~\emph{c}}
        ]
        {\textcolor{blue}{7}~\emph{d}}
    ]

\vspace{0.2cm}

\noindent
The structures (a) and (b) represent the order in which nodes are computed by the top-down and the left-corner parsing strategies, respectively. In the top-down parsing strategy, all the ancestor nodes of \emph{a} must be enumerated before processing \emph{a}. Only when the beam size increases, it is possible to assume various depths and nodes, and thus retain the correct tree structure during sentence processing. On the other hand, in the left-corner parsing strategy, where the mother node is enumerated after its leftmost child, it is not necessary to enumerate the ancestor nodes before processing \emph{a}. Thus, the correct tree structure can be inferred with a small beam size via the left-corner parsing strategy.


Second, we show why inference with a small beam size is comparable to humans. In fact, \citet{jurafskyProbabilisticModelLexical1996} suggested that \emph{full} parallel processing and serial processing are not appropriate for human sentence processing, and instead \emph{partial} parallel processing via beam search that prunes low probability structures is cognitively plausible. From this perspective, we may argue that inference with a small beam size is comparable to humans, and left-corner RNNGs, which can infer the correct tree structure with a small beam size, may be more human-like than top-down RNNGs.

As an anonymous reviewer correctlly pointed out, however, given that \citet{jurafskyProbabilisticModelLexical1996} proposed beam search that only keeps structures with a probability within a multiple of 3.8 to 5.6 of the probability of the most probable structure, the number of structures within such a relative beam could be extremely large, especially if the probabilities of the structures within the beam are similar. In order to address this point, we computed an empirical beam size comparable to humans. Specifically, we calculated the number of structures with a probability more than $1/3.8$ and $1/5.6$ of the probability of the most probable structure, among structures within the word beam which corresponds to the beam defined in \citet{jurafskyProbabilisticModelLexical1996}. The results showed that, even with the largest word beam size of $100$, the average number of structures through derivations within the proposed relative beam turned out to be empirically small: between 3.05 and 4.14 for top-down RNNGs and between 4.08 and 5.68 for left-corner RNNGs. The details of the results are shown in Appendix~\ref{app:empbeam}. We believe that these results do not affect our argument that inference with a small beam size is comparable to humans.

These discussions taken together, we could still argue that left-corner RNNGs, which can infer the correct tree structure with a small beam size, may be more human-like than top-down RNNGs. In addition, given that larger beam sizes make LMs more computationally expensive, these results also suggest that left-corner RNNGs are more efficient.

\section{Limitations and future work}
\label{sec:future}
Interestingly, \citet{wilcoxPredictivePowerNeural2020a} demonstrated that top-down RNNGs underperformed LSTMs in predicting human reading times in English, which appears to be contradictory to our results in Japanese. We would like to suggest that this discrepancy can be attributed to the difference in the languages tested in these papers. In fact, \citet{kuribayashi-etal-2021-lower} have shown that several established results in English cannot be straightforwardly generalized to Japanese.

In addition, \citet{wilcoxPredictivePowerNeural2020a} found that \emph{n}-gram language models outperformed various neural language models, while \citet{merkx-frank-2021-human} observed that Transformers~\cite{Vaswani2017AttentionNeed} outperformed LSTMs in modeling self-paced reading times and N400 brain activities, but not in predicting eye-tracking reading times.

In order to address these limitations, we plan to conduct detailed comparisons between English (Dundee Corpus, \citealp{kennedy2003dundee}) and Japanese (BCCWJ-EyeTrack, \citealp{asaharaReadingTimeAnnotationsBalanced2016}) with RNNGs, incorporating \emph{n}-gram language models and Transformers as baselines in future work.

\section{Conclusion}
\label{sec:conclusion}
In this paper, we investigated whether hierarchical structures make LMs more human-like, and if so, which parsing strategy is most cognitively plausible. Our computational modeling demonstrated that left-corner RNNGs outperformed top-down RNNGs and LSTM, suggesting that hierarchical and left corner architectures are more cognitively plausible than top-down or sequential architectures. Moreover, lower perplexities and higher parsing accuracies of the hierarchical models were strongly correlated with the higher psychometric predictive power, but the correlation did not hold for the sequential model. In addition, left-corner RNNGs may be more human-like than top-down RNNGs in that they can infer the correct tree structure with a small beam size comparable to humans.


\section*{Acknowledgements}
We would like to thank three anonymous reviewers for their insightful suggestions. This work was supported by JSPS KAKENHI Grant Number 19H04990 and the National Institute for Japanese Language and Linguistics (NINJAL) Collaborative Research Project ``Computational Psycholinguistics of Language Processing with Large Corpora.''

\bibliography{anthology,custom}
\bibliographystyle{acl_natbib}

\clearpage
\appendix

{\setlength\textfloatsep{0pt}
\begin{figure}[t]
    \centering
      \includegraphics[width=7.5cm]{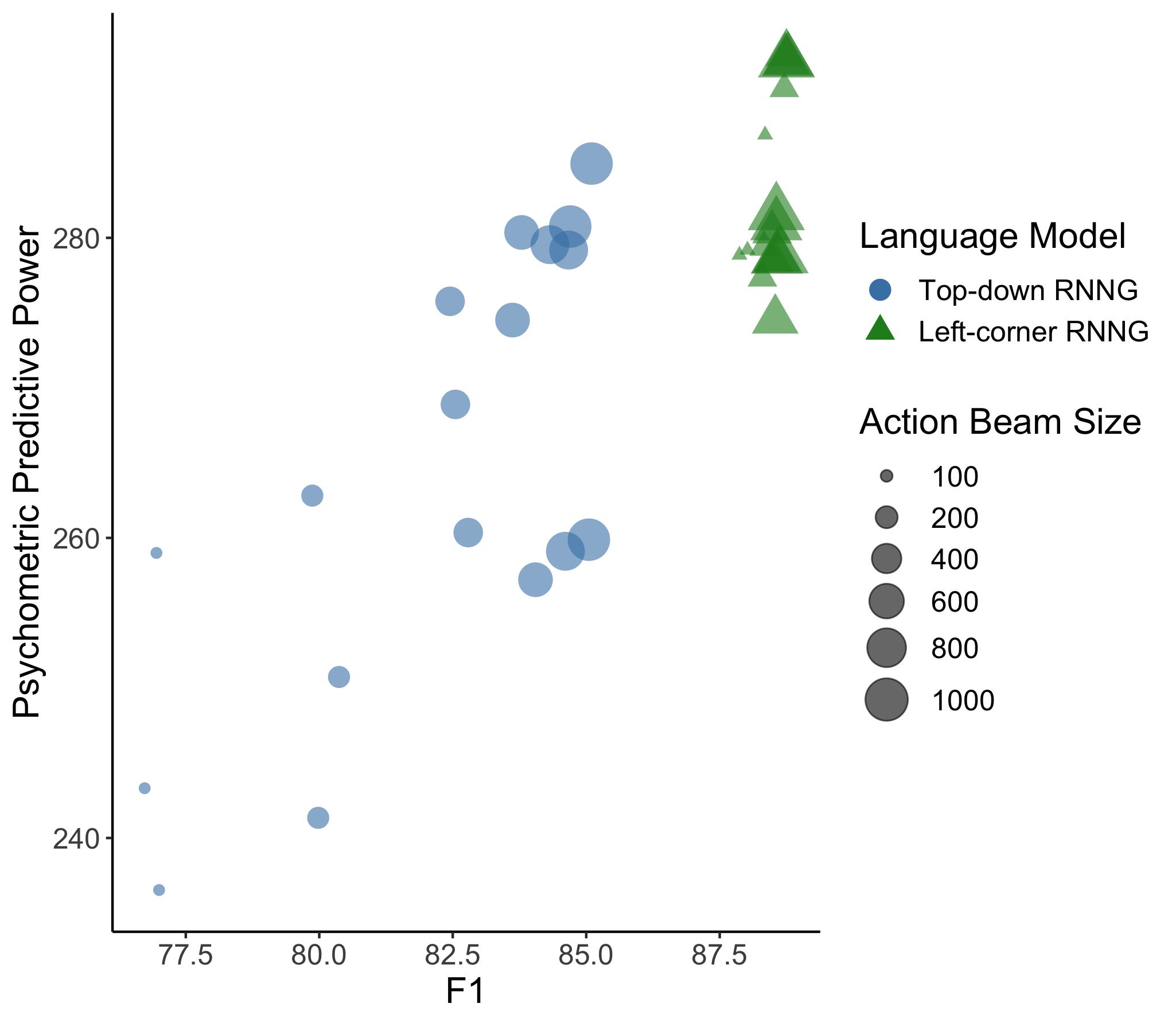}
      \caption{The relationship between parsing accuracy and psychometric predictive power: psychometric predictive power (the vertical axis) is plotted against parsing accuracy (the horizontal axis).}
      \label{fig:f1-dev}
\end{figure}
}

\section{Details of our regression model}
\label{app:variables}
The logarithmic reading time ($\log(\texttt{RT})$) was modeled using the following linear mixed-effects model as a baseline regression model:
\begin{align}
    \nonumber
    \log(\texttt{RT}) &\sim \texttt{length} + \texttt{prev\_length}\\ \nonumber
    &+ \texttt{freq} + \texttt{prev\_freq} + \texttt{is\_first}  \\ \nonumber
    &+ \texttt{is\_last} + \texttt{is\_second\_last} \\ \nonumber
    &+ \texttt{screenN} + \texttt{lineN} + \texttt{segmentN}\\ 
    &+ \texttt{(1|article)} + \texttt{(1|subj)} .
    \label{eq:model}
\end{align}
Table~\ref{tbl:variable} shows descriptions for the factors used in our experiments. We contained predictors and random intercepts that were used in \citet{asaharaReadingTimeAnnotationsBalanced2016}, and added a predictor related to frequency following the previous literature~(e.g., \citealp{frankInsensitivityHumanSentenceProcessing2011}; \citealp{fossumSequentialVsHierarchical2012a}). Frequencies were estimated using the larger National Institute for Japanese Language and Linguistics Web Japanese Corpus (NWJC, \citealp{NWJC}). To capture spillover effects, the length and frequency of the previous segments were also added as a predictor \citep{SMITH2013302}. All numeric factors were centered, and the predictors that were not significant ($p >0.05$) for modeling reading times were excluded. We removed 27 data points that were beyond three standard deviations. This left 12,087 data points as final statistical analysis targets.

{\setlength\textfloatsep{0pt}
\begin{figure}[t]
    \centering
      \includegraphics[width=7.5cm]{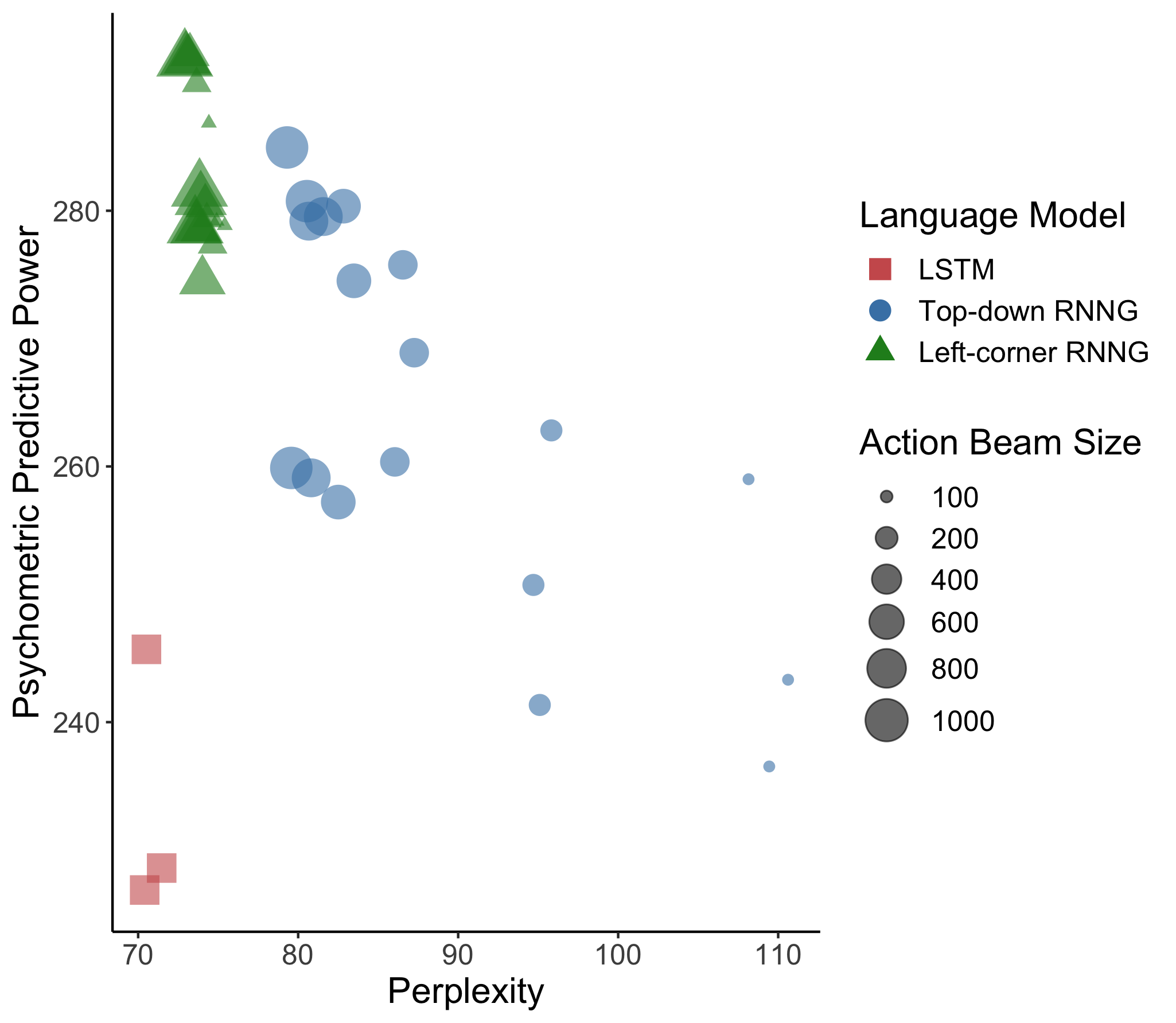}
      \caption{The relationship between perplexity on the NPCMJ test set and psychometric predictive power: psychometric predictive power (the vertical axis) is plotted against perplexity on the NPCMJ test set (the horizontal axis).}
      \label{fig:ppl-dev}
\end{figure}
}

\section{Relationship between parsing accuracy and psychometric predictive power}
\label{app:f1}

The relationship between parsing accuracy and psychometric predictive power is summarized in Figure~\ref{fig:f1-dev}: psychometric predictive power is plotted against parsing accuracy (F1). Just like perplexity, left-corner RNNGs, which achieved higher psychometric predictive power, also achieved the higher parsing accuracy than top-down RNNGs. Overall, the correlation between parsing accuracy and psychometric predictive power of the hierarchical models was robust: the higher parsing accuracy RNNGs have, the higher psychometric predictive power they also have.



\section{Empirical beam size in \citet{jurafskyProbabilisticModelLexical1996}}
\label{app:empbeam}

Table~\ref{tbl:empbeam} shows the average number of structures through derivations within the beam proposed in \citet{jurafskyProbabilisticModelLexical1996}. Specifically, we calculated the number of structures with a probability more than $1/3.8$ and $1/5.6$ of the probability of the most probable structure, among structures within the word beam which corresponds to the beam defined in~\citet{jurafskyProbabilisticModelLexical1996}. The average number of structures within the proposed relative beam turned out to be empirically small.



\section{Hyparameters and available codes}
\label{app:hyper}

Table~\ref{tbl:hypara} shows the hyperparameters of LMs. The code of LSTM we employed \citep{gulordavaColorlessGreenRecurrent2018} is available at \url{https://github.com/facebookresearch/colorlessgreenRNNs}. Surprisals of LSTM were calculated with the code implemented by \citet{van-schijndel-linzen-2018-neural}, which is available at \url{https://github.com/vansky/neural-complexity}. The code of RNNGs we employed \citep{noji-oseki-2021-effective} is also available at \url{https://github.com/aistairc/rnng-pytorch}. We used a byte-pair encoding implemented in sentencepiece \citep{kudoSentencePieceSimpleLanguage2018}, which is available at \url{https://github.com/google/sentencepiece}. We set character coverage to 0.9995, and vocabulary size to 8,000. 


\section{Dataset split ratio}
\label{app:stats}
Sentences in the training data, NPCMJ, are from 14 sources. We used 90\% of sentences in each source as a training set, and 5\% of sentences as a validation set. The remaining 5\% were used as a test set to calculate paring accuracies of RNNGs in Section~\ref{sec:result} and perplexities of LMs in Appendix~\ref{app:pplonNPCMJ}.

\section{Relationship between perplexity on the NPCMJ test set and psychometric predictive power}
\label{app:pplonNPCMJ}
We additionally investigated the relationship between perplexity calculated based on the sentences in the NPCMJ test set and psychometric predictive power. The result is shown in Figure~\ref{fig:ppl-dev}: psychometric predictive power is plotted against perplexity on the NPCMJ test set. Although the perplexities of all LMs were overall lower, there was no substantial difference with the result shown in Figure~\ref{fig:ppl-dev_bccwj}. The difference in the corpus domain may cause the overall difference in perplexity.
\begin{table*}
\centering
\begin{tabular}{llp{10cm}}
\toprule
\textbf{Name}&\textbf{Type}&\textbf{Description}\\
\hline
\texttt{length} &int & Number of characters in the segment\\
\texttt{prev\_length} &int & Number of characters in the previous segment\\
\texttt{freq} &num & Logarithm of the geometric mean of the word frequencies in the segment\\
\texttt{prev\_freq} &num & Logarithm of the geometric mean of the word frequencies in the previous segment\\
\texttt{is\_first} &factor &Whether the segment is the first on a line\\
\texttt{is\_last} &factor &Whether the segment is the last on a line\\
\texttt{is\_second\_last} &factor &Whether the segment is the second to last on a line\\
\texttt{screenN} &int &Screen display order\\
\texttt{lineN} &int &Line display order\\
\texttt{segmentN} &int &Segment display order\\
\texttt{article} &factor &Article ID\\
\texttt{subj} &factor &Participant ID\\
\bottomrule
\end{tabular}
\caption{Descriptions for the factors used in our experiments.}
\label{tbl:variable}
\end{table*}

\begin{table*}
\centering
\small 
\begin{tabular}{llcccccc}
\toprule
\multicolumn{2}{c}{}&\multicolumn{6}{c}{Word beam size}\\
\cmidrule(lr){3-8}
Model&Threshold&$10$ & $20$& $40$& $60$& $80$& $100$\\
\hline
TD & $1/3.8$&1.67 ($\pm$0.0)& 2.14 ($\pm$0.0)& 2.61 ($\pm$0.0)& 2.81 ($\pm$0.1)& 2.95 ($\pm$0.1)& 3.05 ($\pm$0.1)\\
               & $1/5.6$ &1.95 ($\pm$0.0) & 2.62 ($\pm$0.0)& 3.35 ($\pm$0.1)& 3.69 ($\pm$0.1)& 3.95 ($\pm$0.1)& 4.14 ($\pm$0.1)\\
\hline
LC & $1/3.8$&2.58 ($\pm$0.1)& 3.08 ($\pm$0.1)& 3.53 ($\pm$0.2)& 3.83 ($\pm$0.1)& 4.01 ($\pm$0.1)&4.08 ($\pm$0.2)\\
                  &$1/5.6$ &3.14 ($\pm$0.1) & 3.93 ($\pm$0.2)& 4.73 ($\pm$0.2)& 5.21 ($\pm$0.2)& 5.52 ($\pm$0.2)&5.67 ($\pm$0.3)\\
\bottomrule

\end{tabular}
\caption{The average number of structures with a probability more than $1/3.8$ and $1/5.6$ of the probability of the most probable structure, among structures within the word beam which corresponds to the beam defined in~\citet{jurafskyProbabilisticModelLexical1996}. TD and LC indicate top-down and left-corner RNNGs, respectively. Average scores with standard deviations across different random seeds are reported.}
\label{tbl:empbeam}
\end{table*}

\begin{table*}
\centering
\begin{tabular}{llc}
\toprule
LSTM & optimizer & SGD\\
\    &learning rate & 20\\
\    &dropout & 0.2\\
\    &batch size & 64\\
\hline
Top-down RNNGs &optimizer & adam\\
\    &learning rate & 0.001\\
\    &dropout & 0.3\\
\    &batch size & 64\\
\hline
Left-corner RNNGs &optimizer & adam\\
\    &learning rate & 0.001\\
\    &dropout & 0.3\\
\    &batch size & 64\\
\bottomrule
\end{tabular}
\caption{Hyperparameters of LMs.}
\label{tbl:hypara}
\end{table*}


\end{document}